\documentclass[10pt,twocolumn,letterpaper]{article}

\usepackage[pagenumbers]{cvpr} 

\definecolor{cvprblue}{rgb}{0.21,0.49,0.74}
\usepackage[pagebackref,breaklinks,colorlinks,allcolors=cvprblue]{hyperref}

\usepackage{multirow}
\usepackage{booktabs}
\usepackage[table]{xcolor}
\usepackage{amssymb}
\usepackage{adjustbox}
\usepackage{wrapfig}
\usepackage{siunitx}
\usepackage{tcolorbox}
\usepackage{float}
\usepackage{balance}
\usepackage[accsupp]{axessibility} 

\usepackage{algorithm}       
\usepackage{arydshln}
\usepackage{listings}
\newcommand{\listingsfont}{\ttfamily\small}
\usepackage{xcolor}
\definecolor{codegray}{gray}{0.9}
\definecolor{codepurple}{RGB}{163,21,21}
\def\paperID{18} 
\def\confName{CVPR}
\def\confYear{2026}

\title{Image Hashing via Cross-View Code Alignment \\in the Age of Foundation Models}

\author{Ilyass Moummad\thanks{These authors contributed equally.}\\
INRIA, LIRMM, UM\\
\and
Kawtar Zaher\footnotemark[1]\\
INA, INRIA\\
\and
Hervé Goëau\\
CIRAD, AMAP\\
\and
Alexis Joly\\
INRIA, LIRMM, UM\\
}

\begin{document}
\maketitle
\begin{abstract}
Efficient large-scale retrieval requires representations that are both compact and discriminative. Foundation models provide powerful visual and multimodal embeddings, but nearest neighbor search in these high-dimensional spaces is computationally expensive. Hashing offers an efficient alternative by enabling fast Hamming distance search with binary codes, yet existing approaches often rely on complex pipelines, multi-term objectives, designs specialized for a single learning paradigm, and long training times. We introduce \textbf{CroVCA} (\textbf{Cro}ss-\textbf{V}iew \textbf{C}ode \textbf{A}lignment), a simple and unified principle for learning binary codes that remain consistent across semantically aligned views. A single binary cross-entropy loss enforces alignment, while coding-rate maximization serves as an anti-collapse regularizer to promote balanced and diverse codes. To implement this, we design \emph{HashCoder}, a lightweight MLP hashing network with a final batch normalization layer to enforce balanced codes. HashCoder can be used as a probing head on frozen embeddings or to adapt encoders efficiently via LoRA fine-tuning. Across benchmarks, CroVCA achieves state-of-the-art results in just 5 training epochs. At 16 bits, it performs particularly well; for instance, unsupervised hashing on COCO completes in under 2 minutes and supervised hashing on ImageNet100 in about 3 minutes on a single GPU. These results highlight CroVCA's efficiency, adaptability, and broad applicability. Code is available at: \url{https://github.com/ilyassmoummad/cross-view-code-alignment}
\end{abstract}    
\section{Introduction}

Foundation models have reshaped representation learning across vision, language, and multimodal domains~\cite{opportunities, clip, dinov2, dinov3}. Their embeddings capture rich semantic structure and enable applications such as image retrieval, text-to-image search, and recommendation systems. Yet, these embeddings are high-dimensional, making storage and nearest-neighbor search computationally expensive. This motivates the need for compact representations that preserve semantics while enabling fast large-scale retrieval.  

Hashing addresses this challenge by mapping embeddings into binary codes, allowing efficient Hamming distance search with reduced memory and computation~\cite{hashsurvey}. However, learning high-quality hash codes for foundation models remains difficult. Existing methods often rely on multi-stage pipelines or distillation from pretrained embeddings~\cite{hashnet}, and use multi-term objectives to approximate binarization~\cite{ckdh}, enforce alignment~\cite{dhd}, or decorrelate bits~\cite{harr}. These designs complicate optimization, slow convergence, and are typically specialized to a single paradigm, such as unsupervised or supervised hashing~\cite{spq, hashsurvey}.  

This raises a central question:  
\begin{center}
\textit{Can a single, simple framework unify unsupervised and supervised hashing by efficiently leveraging foundation model embeddings?}
\end{center}

We answer this by introducing \textbf{CroVCA} (\textbf{Cro}ss-\textbf{V}iew \textbf{C}ode \textbf{A}lignment), a simple principle for learning binary codes that remain consistent across semantically aligned views. Depending on the setting, aligned views may come from data augmentations (unsupervised) or class-consistent samples (supervised). Alignment is enforced via a binary cross-entropy loss, while coding-rate maximization~\cite{simdino, nmce, empssl} prevents collapse and promotes balanced utilization of the Hamming space. This formulation unifies unsupervised and supervised hashing under a single objective, and can naturally extend to cross-modal scenarios, which we leave for future work.  

To realize this framework, we design \textit{HashCoder}, a lightweight MLP hashing network with a final BatchNorm layer to balance bits. HashCoder can be used as a probing head on frozen embeddings or to efficiently adapt encoders through LoRA fine-tuning~\cite{lora}, supporting both dataset-specific adaptation and transfer from large pretraining datasets (e.g., ImageNet-1k~\cite{imagenet}). We provide two variants: a compact MLP for small datasets (e.g., Flickr25K, COCO, ImageNet100) and a larger one for large-scale datasets (e.g., ImageNet-1k).  

Our contributions are as follows:
\begin{itemize}
    \item We propose \textbf{CroVCA} (\textbf{Cro}ss-\textbf{V}iew \textbf{C}ode \textbf{A}lignment), a simple principle that unifies unsupervised and supervised hashing under one objective.  
    \item We introduce \textbf{HashCoder}, a lightweight MLP hashing network with BatchNorm for balanced bit usage, available in small and large variants.  
    \item We demonstrate \textbf{efficient adaptation} of foundation models via probing and LoRA fine-tuning, enabling dataset-specific and transfer learning scenarios.  
    \item We achieve \textbf{state-of-the-art retrieval performance} with minimal cost—for example, unsupervised hashing on COCO in under 2 minutes and supervised hashing on ImageNet100 in about 3 minutes, trained for only 5 epochs on a single GPU.  
\end{itemize}

\section{Related Works}

\paragraph{Foundation models.} Large-scale pretrained encoders provide versatile visual and multimodal embeddings that serve as backbones for several tasks~\cite{opportunities}. In vision, DINOv3~\cite{dinov3} achieves strong performance on downstream tasks including classification, retrieval, segmentation, and depth estimation. In multimodal settings, text-image models \cite{clip, dfn, blip2} enable zero-shot image classification, captioning, and text-to-image retrieval. These embeddings can be used directly off-the-shelf, through probing with shallow networks or adapted via fine-tuning and parameter-efficient methods such as LoRA~\cite{lora}. Their strong semantic structure makes them attractive candidates for hashing in large-scale retrieval.

\paragraph{Hashing.} Early hashing approaches combined handcrafted features  (e.g., raw pixels, color histograms, edge descriptors) with simple hash functions such as random projections~\cite{lindenstrauss}, PCA~\cite{pcahashing}, or iterative quantization~\cite{itq}. With the rise of pretrained neural networks~\cite{imagenet, alexnet, vgg}, deep hashing replaced handcrafted features with learned embeddings. Initial approaches applied classical hashing to these features~\cite{hashnet}, while subsequent works directly optimized binarization objectives using $\tanh$ or sigmoid relaxations, or straight-through estimators to backpropagate through non-differentiable operations~\cite{hashsurvey}. Supervised hashing often uses class labels to sample triplets or construct class-specific hash centers~\cite{dedaha, dtq, orthohash, csq}, whereas unsupervised methods preserve instance-level consistency, sometimes via knowledge distillation from pretrained encoders~\cite{hashsurvey, vit2hash, fsch, harr} or adversarial regularization~\cite{hashgan}. Self-supervised learning principles such as contrastive objectives~\cite{fsch, cibhash, ctmih, spq}, masked patch modeling~\cite{ctmih}, and entropy maximization~\cite{bihalfnet} have also been applied to unsupervised hashing. Cross-modal hashing extends these ideas to align codes across modalities such as image–text~\cite{ckdh, jdsh, ssah}. A common challenge across these methods is \emph{collapse}, where binary codes degenerate to low-variance solutions; prior work mitigates this with entropy-based regularizers or decorrelation constraints~\cite{bihalfnet, orthohash}. Despite these advances, most methods remain paradigm-specific, and rely on multi-term objectives that are complex and difficult to optimize.

\paragraph{Our distinction.}  
Unlike prior methods that design separate objectives for different paradigms or rely on complex multi-term losses, we propose \textbf{CroVCA} (\textbf{Cro}ss-\textbf{V}iew \textbf{C}ode \textbf{A}lignment), a simple principle for learning binary codes. By aligning semantically consistent views with a binary cross-entropy loss and regularizing with coding-rate maximization~\cite{simdino, nmce, empssl}, our approach avoids code collapse while promoting balanced, high-entropy codes. This formulation unifies unsupervised and supervised hashing within a single objective, removing the need for paradigm-specific designs. While the same principle can naturally extend to cross-modal settings, our focus in this work is on supervised and unsupervised hashing.
\section{Method}

\begin{figure*}[ht] 
  \centering
  \includegraphics[width=0.8\linewidth]{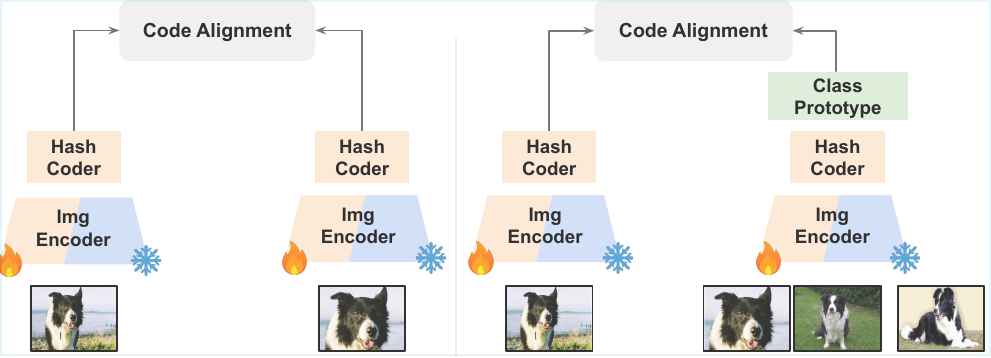}
  \caption{Cross-view code alignment for different hashing setups: unsupervised (left), and supervised (right). Encoders are either frozen or fine-tuned via LoRA.}
  \label{fig:cvca}
\end{figure*}


We introduce \textbf{CroVCA} (\textbf{Cro}ss-\textbf{V}iew \textbf{C}ode \textbf{A}lignment), a unified, information-theoretic framework for learning compact binary codes on top of foundation model embeddings. Figure~\ref{fig:cvca} summarizes the idea across unsupervised and supervised settings. We first introduce the problem statement and the HashCoder module, then derive a principled training objective that balances \emph{alignment} (agreement across views) and \emph{diversity} (code utilization).

\subsection{Notation and problem statement}
Let $\mathcal{X}$ be the input space (images, text, or both) and $b$ the target hash length. The goal is to learn a mapping
\[
\phi:\mathcal{X}\to\{0,1\}^b
\]
that preserves semantic similarity, i.e., semantically related inputs should have small Hamming distance.

For each input $x \in \mathcal{X}$ we construct a \emph{paired} example $(x^{(1)},x^{(2)})$ where the pairing depends on the setting: 
unsupervised, where two augmentations of the same input are used; 
and supervised, where $x^{(2)}$ is a class-representative (prototype or batch-mean) of $x^{(1)}$'s class.

Let $y^{(1)},y^{(2)} \in \{0,1\}^b$ be the observed binary codes for the two views, modeled as realizations of underlying random variables $Y^{(1)}$ and $Y^{(2)}$. The desiderata are: (i) \textbf{alignment}: $d_H(y^{(1)},y^{(2)})$ is small for paired views; and (ii) \textbf{diversity}: the marginal distribution of $Y$ should be high-entropy, ensuring balanced and decorrelated bits.

\subsection{HashCoder: Hashing Network}
We implement $\phi$ as a pretrained encoder $f$ (frozen or adapted via LoRA) followed by a lightweight MLP $g$, called \emph{HashCoder}, that outputs per-bit logits. For a view $x^{(v)}$, $v\in\{1,2\}$:
\begin{align}
h^{(v)} &= f(x^{(v)}) \in \mathbb{R}^d, &\text{(backbone embedding)} \\
z^{(v)} &= g(h^{(v)}) \in \mathbb{R}^b, &\text{(logits)}\\
p^{(v)} &= \sigma\big(z^{(v)}\big)\in[0,1]^b, &\text{(bit probabilities)}\\
y^{(v)} &= \mathbf{1}\{p^{(v)}\ge 0.5\}\in\{0,1\}^b, &\text{(binary code)}
\end{align}
where $\sigma$ is the elementwise sigmoid. 

\begin{figure}
  \centering
  \includegraphics[width=.5\columnwidth]{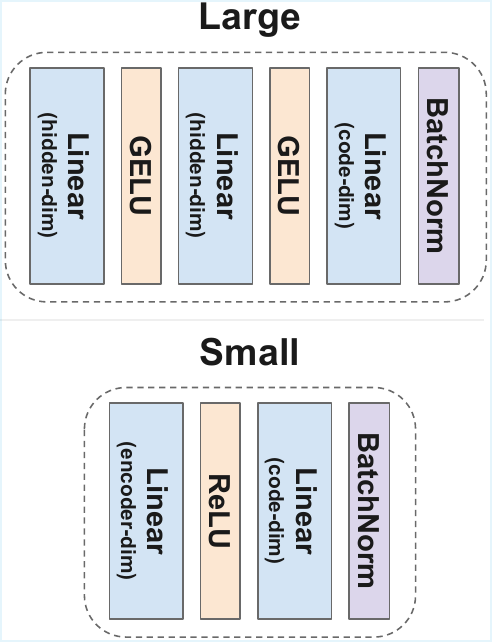}
  \caption{HashCoder design}
  \label{fig:hashcoder_design}
\end{figure}

\paragraph{Architecture and design choices.} HashCoder is a compact MLP inspired by SSL projection heads~\cite{sslcookbook}. We use two variants: (i) a \emph{large} 3-layer MLP for large datasets, and (ii) a \emph{small} 2-layer MLP for lightweight adaptation. Both variants include a final batch normalization layer, implicitly balancing bit usage, following OrthoHash~\cite{orthohash}. Figure~\ref{fig:hashcoder_design} shows the design.

\paragraph{Training dynamics.}  
For each paired example, one branch is binarized to serve as the \emph{teacher} ($y^{(1)}$), while the other branch remains soft ($p^{(2)}$) and acts as the \emph{student}. Gradients are stopped on the teacher, so only the student branch is updated. The roles are swapped symmetrically across views, providing discrete supervision without backpropagating through the hard threshold and thus avoiding the need for straight-through estimators~\cite{ste}.

\subsection{Information-theoretic objective and tractable surrogates}

We aim to increase the mutual information between codes of paired views:
\[
I(Y^{(1)};Y^{(2)}) = H(Y^{(1)}) - H(Y^{(1)}\mid Y^{(2)}),
\]
which decomposes naturally into \textbf{alignment} (small conditional entropy) and \textbf{diversity} (large marginal entropy). Both terms are intractable for discrete high-dimensional codes; we derive tractable surrogates.

\paragraph{Conditional entropy $\to$ binary cross-entropy (alignment).}  
Let $P(Y^{(1)}\!\mid\!Y^{(2)})$ denote the true conditional distribution and $Q(Y^{(1)}\!\mid\!Y^{(2)})$ a surrogate model. By the standard cross-entropy decomposition:
\begin{align*}
    H(Y^{(1)}\mid Y^{(2)}) &= \mathbb{E}[-\log Q(Y^{(1)}\mid Y^{(2)})] - \mathrm{KL}(P\;\|\;Q) \\
    &\le \mathbb{E}[-\log Q(Y^{(1)}\mid Y^{(2)})],
\end{align*}
so the conditional entropy is upper-bounded by the expected negative log-likelihood of the surrogate.

We choose $Q$ to be an elementwise-independent Bernoulli distribution parameterized by the soft outputs $p^{(2)}$ of the other branch:
\[
Q(Y^{(1)} = y^{(1)} \mid p^{(2)}) = \prod_{j=1}^b (p^{(2)}_j)^{y^{(1)}_j} (1-p^{(2)}_j)^{1-y^{(1)}_j}.
\]

The negative log-likelihood under this distribution gives exactly the binary cross-entropy (BCE) between the teacher code $y^{(1)}$ and the student probabilities $p^{(2)}$:
\begin{align*}
\mathrm{BCE}(y^{(1)},p^{(2)}) \;=\; - \sum_{j=1}^b \Big[ &y^{(1)}_j \log p^{(2)}_j \\ + &(1-y^{(1)}_j) \log (1-p^{(2)}_j) \Big].
\end{align*}

Averaging over the batch approximates the expected negative log-likelihood. Symmetrizing across both views yields the alignment loss:
\[
\mathcal{L}_{\mathrm{align}} = \frac{1}{2} \Big[ \mathrm{BCE}(y^{(1)},p^{(2)}) + \mathrm{BCE}(y^{(2)},p^{(1)}) \Big].
\]

Minimizing $\mathcal{L}_{\mathrm{align}}$ therefore reduces an upper bound on the conditional entropy $H(Y^{(1)} \mid Y^{(2)})$, providing a principled surrogate for alignment.

\paragraph{Marginal entropy $\to$ coding-rate surrogate (diversity).}  
The marginal entropy,
\[
H(Y) = - \sum_{y\in\{0,1\}^b} P(Y=y) \log P(Y=y),
\]
promotes balanced, decorrelated bits. Direct computation is infeasible. We define a continuous surrogate using the pre-threshold logits $z$:
\[
v_i = \frac{z_i}{\|z_i\|_2},\quad
C = \frac{1}{B}\sum_{i=1}^B v_i v_i^\top.
\]
Modeling $v_i$ as zero-mean Gaussian, the differential entropy is $h(v) = \frac{1}{2}\log\det(2\pi e \Sigma)$ ~\cite{ma2007}. Maximizing $\log\det\Sigma$ spreads the vectors along independent directions, increasing diversity after thresholding. The numerically stable \emph{coding rate} surrogate is
\[
R(C) = \frac{1}{2}\log\det\Big(I + \frac{d}{B} C\Big),\quad
\mathcal{L}_{\mathrm{div}} = -R(C).
\]

The overall hashing objective is
\[
\mathcal{L}_{\mathrm{hash}} = \mathcal{L}_{\mathrm{align}} + \lambda \mathcal{L}_{\mathrm{div}}, 
\]
with $\lambda>0$ controlling the trade-off. Minimizing this loss simultaneously reduces an upper bound on conditional entropy and increases a differentiable surrogate for marginal entropy, yielding a principled framework for unsupervised and supervised.

\begin{algorithm}
\begin{lstlisting}[basicstyle=\listingsfont\fontsize{7}{8}\selectfont]
# f: encoder (feature extractor)
# g: MLP (HashCoder)
# BCE: BCEWithLogits loss
# lambda: regularization coefficient
# eps: coding rate hyperparameter

# --- MCR regularization function ---
def MCR(X, eps):                     # coding rate
    X = normalize(X, dim=1)          # l2-normalize
    cov = transpose(X) @ X           # covariance matrix
    p = X.shape[1]
    m = X.shape[0]
    I = identity(p)
    scalar = p / (m * eps)
    return -(logdet(I + scalar * cov)) * (p+m)/(p*m)

# --- Training loop ---
for x in loader:
    x1, x2 = aug(x), aug(x)          # generate views
    z1, z2 = f(x1), f(x2)            # encode features
    p1, p2 = g(z1), g(z2)            # project features
    c1, c2 = (p1>0), (p2>0)          # produce codes 
    # code alignment
    L_align = BCE(p1, c2)/2 + BCE(p2, c1)/2
    # diversity regularization
    L_div = MCR(p1, eps)/2 + MCR(p2, eps)/2
    # final loss
    L = L_align + lambda * L_div
    L.backward()                     # backpropagate
    update(f, g)                     # optimizer step
\end{lstlisting}
\caption{Pseudocode of PyTorch training with cross-view code alignment}
\label{alg:bce_mcr_clean}
\end{algorithm}

\section{Experiments}

We evaluate \textbf{CroVCA} on standard image retrieval benchmarks, focusing on \textbf{supervised} and \textbf{unsupervised hashing}. All experiments use either \textbf{HashCoder probing} or \textbf{LoRA fine-tuning}, with models trained for \textbf{5 epochs}. For retrieval, we compute \textbf{asymmetric Hamming distance}~\cite{asymham} between query logits and database codes. Following standard practice, retrieval performance is reported using mean Average Precision (mAP) at typical cutoffs: mAP@1,000 for CIFAR10, ImageNet100, and ImageNet-1k, and mAP@5,000 for FLICKR25K, COCO, and NUS-WIDE. Implementation details are in Table~\ref{tab:dataset_stats}.

\begin{table}[]
\centering
\footnotesize
\caption{Dataset statistics for image retrieval tasks.}
\label{tab:dataset_stats}
\begin{tabular}{lccccc}
\toprule
\textbf{Dataset} & \textbf{Train} & \textbf{Database} & \textbf{Query} & \textbf{Eval Metric} \\
\midrule
Flickr25K & 4,000 & 20,000 & 1,000 & mAP@5k \\
NUS-WIDE-21 & 10,500 & 193,734 & 2,000 & mAP@5k \\
COCO & 10,000 & 117,218 & 5,000 & mAP@5k \\
CIFAR10 & 50,000 & 50,000 & 10,000 & mAP@1k \\
ImageNet100 & 13,000 & 128,503 & 5,000 & mAP@1k \\
ImageNet1K & 1,281,167 & 45,000 & 5,000 & mAP@1k \\
\bottomrule
\end{tabular}
\end{table}

We employed several vision transformer–based backbones in our experiments. 
Table~\ref{tab:models} summarizes the model families, their variants, and pretraining datasets.

\begin{table}[]
\centering
\footnotesize
\caption{Backbone models used in our experiments, with variants and pretraining datasets.}
\label{tab:models}
\resizebox{1.\columnwidth}{!}{%
\begin{tabular}{lll}
\toprule
\textbf{Model Family / Reference} & \textbf{Variants} & \textbf{Training Dataset} \\
\midrule
SimDINOv2 & ViT-Base & IN1k \\
DINOv2 & ViT-Base & LVD-142M \\
DINOv3 & CNX-S, ViT-B, ViT-L & LVD-1689M \\
DFN & ViT-Base, ViT-Large & DFN-2B \\
SWAG & ViT-Base & IG-3.6B $\to$ IN1k \\
DeiT & ViT-Base & IN1k \\
\bottomrule
\end{tabular}
}
\end{table}

Table~\ref{tab:hyperparams} summarizes the hyperparameter settings used across different training protocols.


\begin{table}[h]
\centering
\footnotesize
\caption{Hyperparameter settings for image training protocols.}
\label{tab:hyperparams}
\begin{tabular}{lcc}
\toprule
\textbf{Hyperparameter} & \textbf{Small Datasets} & \textbf{ImageNet-1k} \\
\midrule
\multicolumn{3}{l}{\textbf{Model}} \\ 
\midrule
Lambda & 0.1 & 0.1 \\
LoRA rank & 16 & 16 \\
LoRA dropout & 0.1 & 0.1 \\
HashCoder layers & 2 & 3 \\
\midrule
\multicolumn{3}{l}{\textbf{Data}} \\ 
\midrule
Image size & 224 & 224 \\
Crop size & 40\% & 40\% \\
\# of views & 2 & 2 \\
\midrule
\multicolumn{3}{l}{\textbf{Optimization}} \\ 
\midrule
Batch size & 256 & 256 \\
Optimizer & AdamW & AdamW \\
Learning rate & $1\times10^{-3}$ & $1\times10^{-4}$ \\
Weight decay & $1\times10^{-2}$ & $1\times10^{-4}$ \\
Epochs & 5 & 5 \\
\bottomrule
\end{tabular}
\end{table}

\subsection{Task-Specific Fine-Tuning}

\textbf{Question 4.1.1:} Can foundation model embeddings be efficiently adapted into compact binary codes with lightweight unsupervised fine-tuning?  

\textbf{Experiment 4.1.1:} We train HashCoder by fine-tuning foundation model embeddings with LoRA at 16, 32, and 64 bits using unsupervised hashing. Retrieval performance is compared with state-of-the-art unsupervised hashing results in Table~\ref{tab:loraft_unsupervisedhashing}. To assess the contribution of key components, we additionally ablate the SWAG backbone under 16-bit unsupervised hashing by removing (i) the final-layer BatchNorm and (ii) the coding-rate regularization term in Table~\ref{tab:swag_ablation_16}.

\begin{table*}[]
    \centering
    \caption{Unsupervised hashing of pre-trained foundation models in comparison to state-of-the-art. \textbf{Best} and \underline{second best} results are highlighted.}
    \label{tab:loraft_unsupervisedhashing}
    \begin{adjustbox}{width=\textwidth}
    \begin{tabular}{ccccccccccccccccccccc}
    \toprule
    \textbf{Model} 
    & \multicolumn{4}{c}{\textbf{CIFAR10}} 
    & \multicolumn{4}{c}{\textbf{COCO}} 
    & \multicolumn{4}{c}{\textbf{FLICKR25K}} 
    & \multicolumn{4}{c}{\textbf{NUS-WIDE}} 
    & \multicolumn{4}{c}{\textbf{ImageNet100}} \\
    \cmidrule(lr){2-5} \cmidrule(lr){6-9} \cmidrule(lr){10-13} \cmidrule(lr){14-17} \cmidrule(lr){18-21}
    & Orig & 16 & 32 & 64 & Orig & 16 & 32 & 64 & Orig & 16 & 32 & 64 & Orig & 16 & 32 & 64 & Orig & 16 & 32 & 64 \\
    \midrule
    
    \rowcolor{gray!25}
    \multicolumn{21}{l}{\textbf{SOTA}} \\

    \rowcolor{gray!10}
    \textbf{IPHASH\textsuperscript{\citep{vit2hash}}} & - & 94.2 & 95.1 & 95.8 & - & 82.6 & 87.5 & \underline{89.4} & - & - & - & - &  - & 79.7 & 81.6 & 82.6 & - & - & - & - \\
    \rowcolor{gray!10}
    \textbf{HARR*\textsuperscript{\citep{harr}}} & - & - & - & - & - & 74.8 & 78.9 & 81.6 & - & 81.8 & 83.0 & 83.8 & - & 80.7 & 82.6 & 84.1 & - & - & - & - \\
    \rowcolor{gray!10}
    \textbf{FSCH\textsuperscript{\citep{fsch}}} & - & 87.6 & 91.2 & 92.6 & - & 76.0 & 78.7 & 79.9 & - & 81.5 & \textbf{83.8} & \textbf{84.9} & - & \underline{81.2} & \textbf{83.2} & \textbf{84.4} & - & - & - & - \\
    \rowcolor{gray!10}
    \textbf{CTMIH\textsuperscript{\citep{ctmih}}} & - & - & - & - & - & 80.9 & 83.4 & 84.6 & - & - & - & - & - & 79.5 & 81.6 & 82.6 & - & 82.0 & 86.0 & 86.9 \\
    
    \midrule[1pt]
    \rowcolor{blue!25}
    \multicolumn{21}{l}{\textbf{CroVCA (ours)}} \\
    \rowcolor{blue!10}
    \multicolumn{21}{l}{\textbf{SimDINOv2} \ \footnotesize{(IN-1k)}} \\
    \shortstack{ViT-B} 
    & 89.6 & 95.9 & 95.1 & 93.8 
    & 87.4 & 85.4 & 87.0 & 87.8 
    & 81.1 & 78.3 & 78.0 & 75.6 
    & 84.3 & \textbf{81.8} & 81.8 & 82.4 
    & 84.1 & 79.6 & 81.8 & 83.8 \\
    \rowcolor{blue!10}
    \multicolumn{21}{l}{\textbf{DINOv2} \ \footnotesize{(LVD-142M)}} \\
    \shortstack{ViT-B} 
    & 95.4 & \textbf{98.6} & \textbf{98.7} & \textbf{97.9} 
    & 88.3 & \textbf{87.5} & \textbf{89.2} & 89.0 
    & 76.3 & 69.1 & 69.1 & 68.2 
    & 79.8 & 75.7 & 77.4 & 77.3 
    & 88.2 & 87.1 & 88.5 & 89.2 \\
    \rowcolor{blue!10}
    \multicolumn{21}{l}{\textbf{DINOv3} \ \footnotesize{(LVD-1689M)}} \\
    \shortstack{ViT-S}
    & 86.9 & 93.8 & 92.5 & 90.6 
    & 82.7 & 79.5 & 82.3 & 82.4 
    & 72.9 & 68.2 & 67.5 & 66.7 
    & 81.4 & 80.8 & 81.2 & 80.1 
    & 77.7 & 60.8 & 72.1 & 74.8 \\
    \shortstack{ViT-B} 
    & 93.6 & \underline{97.7} & \underline{97.7} & \underline{96.7} 
    & 86.6 & \underline{86.7} & 88.8 & 89.1 
    & 73.0 & 64.2 & 65.3 & 65.4 
    & 80.7 & 76.7 & 78.9 & 78.9 
    & 85.9 & 83.5 & 86.1 & 88.9 \\
    \rowcolor{blue!10}
    \multicolumn{21}{l}{\textbf{DFN} \ \footnotesize{(DFN-2B)}} \\
    \shortstack{ViT-B} 
    & 94.2 & 93.0 & 93.2 & 93.6 
    & 87.0 & 83.4 & 87.5 & 89.2 
    & 83.1 & \textbf{83.3} & \underline{82.4} & \underline{81.9} 
    & 84.1 & 80.4 & \underline{83.1} & \underline{83.0} 
    & 81.1 & 61.9 & 73.9 & 78.8 \\
    \rowcolor{blue!10}
    \multicolumn{21}{l}{\textbf{SWAG} \ \footnotesize{(IG-3.6B $\to$ IN1k)}} \\
    \shortstack{ViT-B} 
    & 89.5 & 92.8 & 91.7 & 90.5 
    & 88.6 & 84.7 & \underline{89.0} & \textbf{90.1} 
    & 79.2 & \underline{82.2} & 80.7 & 78.5 
    & 82.5 & 80.6 & 82.2 & 82.6
    & 94.3 & \textbf{91.9} & \textbf{93.3} & \textbf{94.3} \\
    \rowcolor{blue!10}
    \multicolumn{21}{l}{\textbf{DeiT} \ \footnotesize{(IN-1k)}} \\
    \shortstack{ViT-B} 
    & 84.3 & 89.8 & 90.2 & 88.9 
    & 84.9 & 82.1 & 84.8 & 85.4 
    & 77.9 & 80.4 & 80.0 & 78.7 
    & 82.3 & 80.1 & 81.8 & 82.4 
    & 90.9 & \underline{91.5} & \underline{92.9} & \underline{93.7} \\
    \bottomrule
    \end{tabular}
    \end{adjustbox}
\end{table*}

\textbf{Findings 4.1.1:} LoRA fine-tuning with cross-view code alignment consistently matches or surpasses prior unsupervised hashing methods across datasets and bit lengths. Even after only 5 epochs, HashCoder produces competitive or state-of-the-art retrieval performance. 
Component ablations show that removing BatchNorm leads to consistent performance degradation, whereas removing the coding-rate regularization results in representation collapse, demonstrating that this regularizer is essential for learning stable and informative hash codes.

\begin{tcolorbox}
[colback=gray!2,
colframe=gray!15,
coltitle=black,
title=\textbf{Takeaway 4.1.1}]
Cross-view code alignment efficiently learns compact binary codes via lightweight LoRA fine-tuning in an unsupervised manner, while preserving class-level semantic structure.
\end{tcolorbox}

\begin{table}[!ht]
\centering
\caption{Ablation study of the SWAG encoder for \textbf{16-bit} unsupervised hashing. We report mAP@all and performance change relative to the full SWAG model.}
\label{tab:swag_ablation_16}
\begin{adjustbox}{max width=\columnwidth}
\begin{tabular}{lccccc}
\toprule
\textbf{Model (16-bit)} 
& \textbf{CIFAR10} 
& \textbf{COCO} 
& \textbf{FLICKR25K} 
& \textbf{NUS-WIDE} 
& \textbf{ImageNet100} \\
\midrule
\textbf{SWAG (full)} 
& 92.80 & 84.70 & 82.20 & 80.60 & 91.90 \\
\midrule
\textbf{w/o BatchNorm} 
& 85.27 & 82.86 & 76.50 & 80.96 & 86.83 \\
$\Delta$ 
& {\color{red}-7.53} & {\color{red}-1.84} & {\color{red}-5.70} & {\color{green}+0.36} & {\color{red}-5.07} \\
\midrule
\textbf{w/o Coding Rate} 
& 10.58 & 34.88 & 53.81 & 36.87 & 1.90 \\
$\Delta$ 
& {\color{red}-82.22} & {\color{red}-49.82} & {\color{red}-28.39} & {\color{red}-43.73} & {\color{red}-90.00} \\
\bottomrule
\end{tabular}
\end{adjustbox}
\end{table}

\textbf{Question 4.1.2:} How well is semantic structure preserved when embeddings are compressed into very short binary codes via unsupervised hashing?

\begin{figure}[]
  \centering
  \begin{minipage}{0.48\textwidth}
    \centering
    \includegraphics[width=\textwidth]{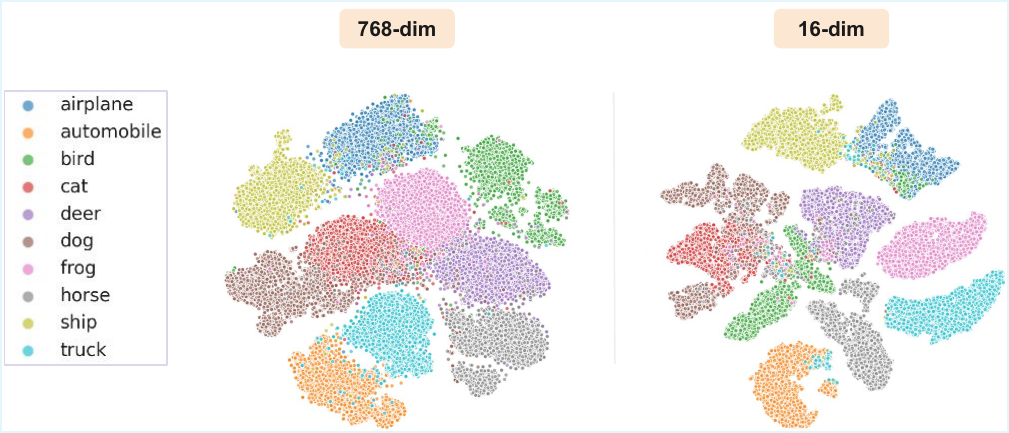}
    \caption{t-SNE of CIFAR10 embeddings: original 768-dim (left) vs. 16-dim HashCoder (right).}
    \label{fig:tsne_cifar10}
  \end{minipage}\hfill
  \begin{minipage}{0.48\textwidth}
    \centering
    \includegraphics[width=\textwidth]{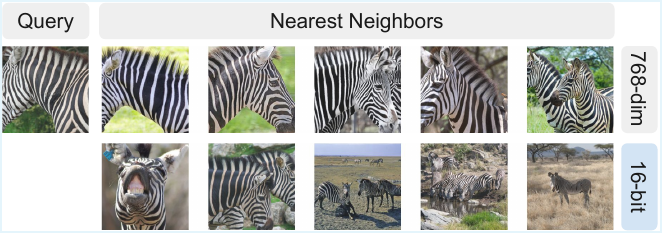}
    \caption{Nearest neighbors of a zebra using SimDINOv2 features vs. HashCoder's 16-bit codes.}
    \label{fig:code_diversity}
  \end{minipage}
\end{figure}

\textbf{Experiment 4.1.2:} To evaluate semantic preservation, we visualize 16-bit HashCoder embeddings on CIFAR10 using t-SNE and compare them with the original 768-dimensional embeddings (Figure~\ref{fig:tsne_cifar10}). Additionally, we perform nearest neighbor retrieval on ImageNet100: for a zebra query, we compare results using the original 768-dim features versus the 16-bit HashCoder codes (Figure~\ref{fig:code_diversity}). In both cases, we start from pretrained SimDINOv2 features and train HashCoder on top via unsupervised hashing using LoRA finetuning of the backbone.

\textbf{Findings 4.1.2:} Even with over 40× dimensionality reduction, the 16-bit HashCoder codes preserve the overall class structure. On CIFAR10, t-SNE shows clearly separable clusters corresponding to different categories. On ImageNet100, nearest neighbor retrieval indicates that HashCoder captures semantic features: for a zebra query, it retrieves diverse zebra images, whereas the original embeddings mostly select visually near-identical images. This demonstrates that HashCoder’s binary representations retain meaningful semantic organization despite extreme compression.

\begin{tcolorbox}
[colback=gray!2,
colframe=gray!15,
coltitle=black,
title=\textbf{Takeaway 4.1.2}]
Unsupervised cross-view code alignment compresses embeddings into 16-bit codes while preserving class-level semantics, enabling retrieval of diverse and semantically meaningful images.
\end{tcolorbox}

\textbf{Question 4.1.3:} How does cross-view code alignment perform compared to state-of-the-art supervised methods for compact code learning?

\begin{table}[]
    \centering
    \caption{Supervised hashing results on IN100 and IN1k datasets. Best and second best results are highlighted.}
    \label{tab:sh_full}
    \resizebox{1.\columnwidth}{!}{%
    \begin{tabular}{l c ccc ccc}
    \toprule
    \textbf{Method} & \textbf{Epochs} 
    & \multicolumn{3}{c}{\textbf{IN100}} 
    & \multicolumn{3}{c}{\textbf{IN1k}} \\
    \cmidrule(lr){3-5} \cmidrule(lr){6-8}
    & & 16 & 32 & 64 & 16 & 32 & 64 \\
    \midrule
    \rowcolor{gray!25}
    \textbf{SOTA} & & & & & & & \\
    \rowcolor{gray!10}
    CSQ\textsuperscript{\citep{csq}} & 90 & 83.7 & 87.5 & 88.7 & 50.4 & 60.6 & 60.9 \\
    \rowcolor{gray!10}
    GreedyHash\textsuperscript{\citep{greedyhash}} & 120 & 85.4 & 87.9 & 88.5 & 54.2 & 58.9 & 59.5 \\
    \rowcolor{gray!10}
    OrthoHash\textsuperscript{\citep{orthohash}} & 100 & 86.9 & 88.6 & 89.9 & \underline{59.3} & \underline{65.1} & \textbf{67.6} \\
    \rowcolor{gray!10}
    FPPQ\textsuperscript{\citep{fppq}} & 100 / 90 & 89.5 & 91.2 & 91.5 & \textbf{62.0} & \textbf{65.4} & \underline{66.4} \\
    \midrule[1pt]
    \rowcolor{blue!25}
    \textbf{Ours} & & & & & & & \\
    \rowcolor{blue!10}
    DINOv2 & 5 & 90.2 & 91.1 & 92.1 & 58.4 & 64.3 & 65.9 \\
    \rowcolor{blue!10}
    SWAG & 5 & \underline{91.9} & \textbf{93.4} & \textbf{94.4} & 54.8 & 63.3 & 65.9 \\
    \rowcolor{blue!10}
    DeiT & 5 & \textbf{92.5} & \underline{93.0} & \underline{93.5} & 54.6 & 59.7 & 62.2 \\
    \bottomrule
    \end{tabular}%
    }
\end{table}

\textbf{Experiment 4.1.3:} We train HashCoder with class supervision on ImageNet100 and ImageNet-1k, and compare it against FPPQ~\cite{fppq}, the state-of-the-art supervised quantization method, and OrthoHash~\cite{orthohash}, the state-of-the-art supervised hashing method (Table~\ref{tab:sh_full}). All experiments use 16-, 32-, and 64-bit codes.

\textbf{Findings 4.1.3:} Our approach surpasses FPPQ and OrthoHash on ImageNet100 and achieves competitive performance on ImageNet-1k, despite using only 5 training epochs compared to 90–100 for the other methods. This highlights that cross-view code alignment enables efficient supervised compact code learning.

\begin{tcolorbox}
[colback=gray!2,
colframe=gray!15,
coltitle=black,
title=\textbf{Takeaway 4.1.3}]
Cross-view code alignment efficiently produces supervised hash codes while requiring very few training iterations.
\end{tcolorbox}

\textbf{Question 4.1.4:} How does our method perform qualitatively compared to an efficient hashing approach?

\begin{figure*}[]
  \centering
  \includegraphics[width=\linewidth]{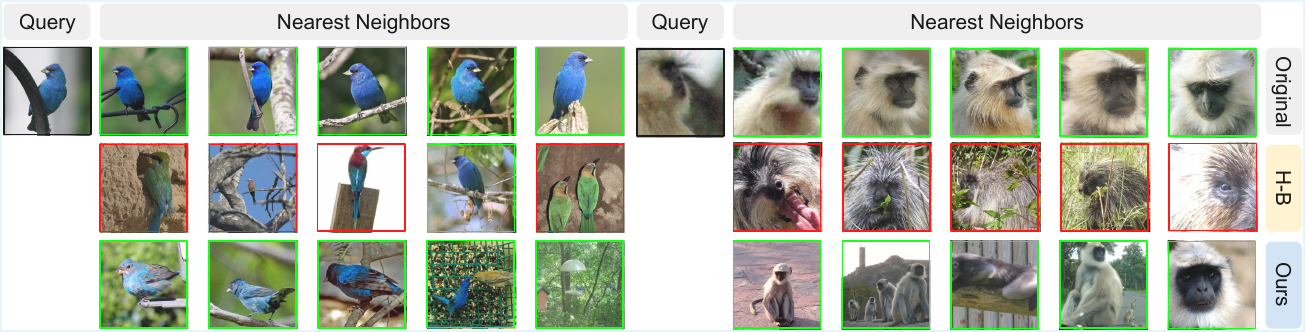}
  \caption{ImageNet100 retrieval results for two queries. 
  Rows: original SimDINOv2 768-dim embeddings; 
  Hashing-Baseline (H-B)~\cite{hashing-baseline}; 
  HashCoder with 16-bit codes trained via unsupervised LoRA fine-tuning.}
  \label{fig:retrieval_simdino}
\end{figure*}

\begin{figure}[] 
  \centering
  \includegraphics[width=0.9\linewidth]{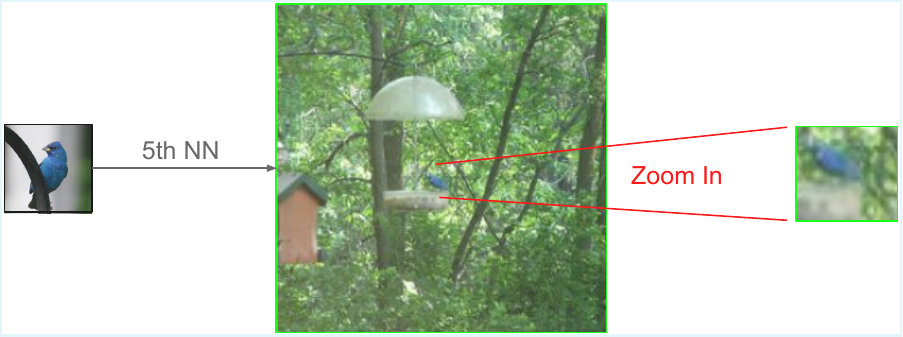}
  \caption{Zoom in on the fifth-nearest neighbor of the query image of a blue bird in ImageNet100 using our 16-bit hashed representations of SimDINOv2.}
  \label{fig:zoom}
\end{figure}

\textbf{Experiment 4.1.4:} We perform nearest-neighbor retrieval on ImageNet100 for two carefully chosen queries of an indigo bird and a grey langur, to highlight fine-grained distinctions and visual ambiguity. We compare three methods: 
(1) cosine similarity on the original 768-dimensional SimDINOv2 embeddings, 
(2) Hashing-Baseline (H-B)~\cite{hashing-baseline}, a fast PCA-based 16-bit hashing method, and 
(3) HashCoder with 16-bit codes trained via LoRA fine-tuning for 5 epochs on a single GPU in 3 minutes (Figure~\ref{fig:retrieval_simdino}).
\\ \indent
\textbf{Findings 4.1.4:} For the indigo bird query, the original embeddings retrieve visually consistent images of the species. Hashing-Baseline~\cite{hashing-baseline}, although efficient, retrieves only one correct image and four unrelated birds, showing a loss of fine-grained semantic detail. In contrast, HashCoder successfully retrieves all five correct and diverse images, including a very small bird (see Figure~\ref{fig:zoom} for a zoom), demonstrating effective preservation of semantic information under 16-bit compression. For the grey langur query, HashCoder retrieves nearest neighbors from the correct class, with different backgrounds, whereas Hashing-Baseline returns visually similar but semantically incorrect animals. These results highlight that cross-view code alignment maintains semantic structure and is robust to visually challenging or ambiguous queries.

\begin{tcolorbox}
[colback=gray!2,
colframe=gray!15,
coltitle=black,
title=\textbf{Takeaway 4.1.4}]
Cross-view code alignment preserves fine-grained class semantics and outperforms fast PCA-based hashing under extreme 16-bit compression, even for ambiguous or visually challenging queries.
\end{tcolorbox}

\subsection{Transferability of ImageNet Fine-Tuning}

\textbf{Question 4.2.1:} Can a single HashCoder trained via cross-view code alignment on a large dataset such as ImageNet-1k generate hash codes that transfer effectively to downstream datasets, reducing the need for per-task retraining?

\textbf{Experiment 4.2.1:} We train HashCoder on ImageNet-1k via LoRA for multiple bit lengths (16, 32, 64) and evaluate the transfer of these hash codes to CIFAR10, FLICKR25K, COCO, NUS-WIDE, and ImageNet100 without any additional training. Both unsupervised and supervised fine-tuning settings are considered (Table~\ref{tab:loraft_transfer}).

\begin{table*}[]
    \centering
    \caption{Transferability of hash codes from ImageNet-1k to downstream datasets.}
    \label{tab:loraft_transfer}
    \resizebox{\linewidth}{!}{%
    \begin{tabular}{ccccccccccccccccccccc}
    \toprule
    \textbf{Method} 
    & \multicolumn{4}{c}{\textbf{CIFAR10}} 
    & \multicolumn{4}{c}{\textbf{COCO}} 
    & \multicolumn{4}{c}{\textbf{FLICKR25K}} 
    & \multicolumn{4}{c}{\textbf{NUS-WIDE}} 
    & \multicolumn{4}{c}{\textbf{ImageNet100}} \\
    \cmidrule(lr){2-5} \cmidrule(lr){6-9} \cmidrule(lr){10-13} \cmidrule(lr){14-17} \cmidrule(lr){18-21}
    & Orig & 16 & 32 & 64 & Orig & 16 & 32 & 64 & Orig & 16 & 32 & 64 & Orig & 16 & 32 & 64 & Orig & 16 & 32 & 64 \\
    \midrule
    \rowcolor{gray!25}
    \multicolumn{21}{l}{\textbf{Unsupervised}} \\
    DINOv2 & 95.4 & 93.2 & 95.0 & 95.8 & 88.3 & 82.8 & 85.8 & 86.6 & 76.3 & 76.5 & 78.9 & 78.2 & 79.8 & 74.0 & 76.0 & 77.6 & 88.2 & 79.9 & 84.0 & 85.6 \\
    SimDINOv2 & 89.6 & 77.3 & 79.5 & 81.3 & 87.4 & 82.5 & 83.3 & 84.9 & 81.1 & 78.0 & 78.2 & 78.3 & 84.3 & 76.0 & 77.4 & 78.7 & 84.1 & 72.8 & 77.8 & 79.6 \\
    \midrule[1pt]
    \rowcolor{gray!25}
    \multicolumn{21}{l}{\textbf{Supervised}} \\
    DINOv2 & 95.4 & 92.9 & 95.2 & 95.7 & 88.3 & 82.2 & 85.8 & 86.7 & 76.3 & 75.4 & 77.0 & 76.7 & 79.8 & 71.6 & 75.2 & 76.4 & 88.2 & 83.0 & 87.8 & 88.3 \\
    SimDINOv2 & 89.6 & 78.8 & 82.1 & 82.7 & 87.4 & 79.9 & 83.2 & 84.0 & 81.1 & 76.3 & 76.5 & 76.7 & 84.3 & 73.6 & 76.2 & 76.8 & 84.1 & 74.6 & 81.0 & 83.0 \\
    \bottomrule
    \end{tabular}
    }
\end{table*}

\textbf{Findings 4.2.1:} HashCoder trained on ImageNet-1k transfers effectively to all downstream datasets, with only minor drops in performance compared to the original continuous representations. This holds for both unsupervised and supervised hashing, showing that cross-view code alignment produces semantically rich codes that generalize well across datasets.

\begin{tcolorbox}
[colback=gray!2,
colframe=gray!15,
coltitle=black,
title=\textbf{Takeaway 4.2.1}]
A single HashCoder trained via cross-view code alignment on a large-scale dataset like ImageNet-1k can generate compact hash codes that transfer effectively to downstream tasks, reducing the need for per-task retraining.
\end{tcolorbox}

\subsection{Transferability of ImageNet-1k Probing}
\label{transfer_probing}

Previous experiments demonstrated that LoRA fine-tuning can efficiently produce low-bit hash codes with competitive retrieval performance in just a few training iterations. We then showed that a HashCoder fine-tuned on a large-scale dataset such as ImageNet exhibits good transferability to new tasks with minor performance degradation. However, these results still rely on fine-tuning. This motivates investigating whether a HashCoder trained once on ImageNet via probing can serve as a general-purpose module that produces compact, transferable embeddings across diverse downstream tasks.

\textbf{Question 4.3.1:} Can a single, general-purpose HashCoder, trained via cross-view code alignment on a frozen foundation model using a large-scale dataset, produce compact codes that preserve semantic structure and transfer effectively across different models and downstream tasks?

\begin{table}[]
\centering
\small 
\caption{Ablation of code dimension with DINOv3 CNX-S using HashCoder probing on IN1k. \textbf{Best} and \underline{second best} results are highlighted.}
\label{tab:abla_code}
\begin{adjustbox}{width=1.\columnwidth}
\begin{tabular}{cccccc}
\toprule
\textbf{Features} 
& \textbf{CIFAR10} 
& \textbf{COCO} 
& \textbf{FLICKR25K} 
& \textbf{NUS-WIDE} 
& \textbf{ImageNet100} \\
\midrule
\rowcolor{gray!25}
\shortstack{\textcolor{gray}{768-dim}} 
& \textcolor{gray}{94.1} & \textcolor{gray}{86.7} & \textcolor{gray}{75.5} & \textcolor{gray}{78.1} & \textcolor{gray}{88.3} \\
\midrule
256-bit & \textbf{93.2} & \textbf{87.3} & 75.8 & \underline{77.2} & \textbf{89.5} \\
128-bit & \underline{92.9} & \underline{86.6} & 76.0 & \textbf{77.3} & \underline{88.4} \\
64-bit  & 92.1 & 85.4 & \underline{77.4} & 76.6 & 87.2 \\
32-bit  & 90.8 & 84.1 & \textbf{78.5} & 74.5 & 84.8 \\
16-bit  & 85.9 & 78.6 & 77.3 & 72.2 & 80.7 \\
\bottomrule
\end{tabular}
\end{adjustbox}
\end{table}

\begin{table}[]
\centering
\small 
\caption{Transfer learning of HashCoder probing on IN1k. Retrieval is evaluated across multiple datasets using original and hashed 256-bit codes.}
\label{tab:transfer_learning}
\resizebox{1.\columnwidth}{!}{%
\begin{tabular}{ccccccccccc}
\toprule
\textbf{Model} 
& \multicolumn{2}{c}{\textbf{CIFAR10}} 
& \multicolumn{2}{c}{\textbf{COCO}} 
& \multicolumn{2}{c}{\textbf{FLICKR25K}} 
& \multicolumn{2}{c}{\textbf{NUS-WIDE}} 
& \multicolumn{2}{c}{\textbf{ImageNet100}} \\
\cmidrule(lr){2-3} \cmidrule(lr){4-5} \cmidrule(lr){6-7} \cmidrule(lr){8-9} \cmidrule(lr){10-11}
& Orig & Code & Orig & Code & Orig & Code & Orig & Code & Orig & Code \\
\midrule
\rowcolor{gray!25}
\multicolumn{11}{l}{\textbf{Random HashCoder}} \\
\rowcolor{gray!10}
\shortstack{ViT-B} 
& 93.6 & 87.4 & 86.6 & 74.5 & 73.0 & 62.2 & 80.7 & 68.7 & 85.9 & 77.7 \\
\midrule[1pt]
\rowcolor{gray!25}
\multicolumn{11}{l}{\textbf{Unsupervised}} \\
ViT-L & 96.9 & 97.4 & 86.9 & 88.3 & 73.3 & 76.1 & 79.8 & 78.3 & 90.2 & 92.3 \\
ViT-B & 93.6 & 94.8 & 86.6 & 88.4 & 73.0 & 78.5 & 80.7 & 80.2 & 85.9 & 87.6 \\
CNX-S & 94.1 & 93.2 & 86.7 & 87.3 & 75.5 & 75.8 & 78.1 & 77.2 & 88.3 & 89.5 \\
\midrule[1pt]
\rowcolor{gray!25}
\multicolumn{11}{l}{\textbf{Supervised}} \\
ViT-L & 96.9 & 97.4 & 86.9 & 87.5 & 73.3 & 76.3 & 79.7 & 78.7 & 90.2 & 93.5 \\
ViT-B & 93.6 & 94.6 & 86.6 & 87.8 & 73.0 & 77.0 & 80.7 & 78.9 & 85.9 & 89.7 \\
CNX-S & 94.1 & 93.1 & 86.7 & 86.6 & 75.5 & 74.6 & 78.1 & 76.6 & 88.3 & 91.3 \\
\bottomrule
\end{tabular}
}
\end{table}

\textbf{Experiment 4.3.1a:} To determine the minimal code length that preserves the semantic richness of the original embeddings, we conduct an ablation study with code lengths ranging from 16 to 256 bits. We use DINOv3’s ConvNext-Small (CNX-S) as a frozen backbone and apply HashCoder probing on ImageNet-1k to generate hash codes of different lengths. These codes are then transferred to downstream datasets (CIFAR10, FLICKR25K, COCO, NUS-WIDE, and ImageNet100) to evaluate retrieval performance. Table~\ref{tab:abla_code} summarizes the results, allowing us to identify code lengths that match or approach the performance of the full embeddings.
\\ \indent
\textbf{Findings 4.3.1a:} Retrieval performance generally improves with longer codes, with 256-bit codes achieving the best balance across datasets, closely matching or surpassing the original embeddings. Shorter codes (16–64 bits) still maintain reasonable performance, demonstrating that compact hash codes preserve substantial semantic information.
\\ \indent
\textbf{Experiment 4.3.1b:} Using the 256-bit codes from the ablation, we perform HashCoder probing on ImageNet-1k with ViT-L, ViT-B, and CNX-S backbones under both unsupervised and supervised settings. The resulting codes are transferred to downstream datasets to evaluate transferability (Table~\ref{tab:transfer_learning}).
\\ \indent
\textbf{Findings 4.3.1b:} Across supervised and unsupervised settings, 256-bit codes obtained via HashCoder probing preserve the semantic information of the original embeddings. Retrieval performance on downstream datasets is comparable to , often better than, full embeddings, demonstrating that HashCoder can generate compact, transferable codes efficiently from a single large-scale dataset.

\begin{tcolorbox}
[colback=gray!2,
colframe=gray!15,
coltitle=black,
title=\textbf{Takeaway 4.3.1}]
Cross-view code alignment enables frozen foundation models to produce compact hash codes that preserve semantics—creating lightweight, transferable hashing networks without any backbone retraining.
\end{tcolorbox}

\section{Conclusion}

We introduced \textbf{CroVCA} (\textbf{Cro}ss-\textbf{V}iew \textbf{C}ode \textbf{A}lignment), a simple and efficient framework for adapting foundation models to hashing. It leverages a lightweight hashing network, \textit{HashCoder}, trained either by probing frozen embeddings or via LoRA fine-tuning, supporting both supervised and unsupervised settings. By aligning views through maximization of mutual information, our method produces compact binary codes that preserve the semantic structure of the original embeddings with minimal training. Even extremely low-bit codes capture meaningful class-level information in a fully unsupervised manner. Future work will extend CroVCA to fine-grained retrieval and explore transferability to new domains.
\section*{Acknowledgment}

This work was supported by the Pl@ntAgroEco project, funded by the "Agence Nationale de la Recherche" (ANR) under the France 2030 program, within the “Agroécologie et Numérique” initiative (reference ANR-22-PEAE-0009). The authors gratefully acknowledge this support.
\balance
{
    \small
    \bibliographystyle{ieeenat_fullname}
    \bibliography{main}
}


\end{document}